\documentclass{article}

\PassOptionsToPackage{numbers, compress}{natbib}
\usepackage{graphicx}
\usepackage{soul} %comment it when not using highlight
\usepackage{setspace}
\usepackage{amsmath}

\usepackage[preprint]{neurips_2025}

\usepackage[utf8]{inputenc} % allow utf-8 input
\usepackage[T1]{fontenc}    % use 8-bit T1 fonts
\usepackage{hyperref}       % hyperlinks
\usepackage{url}            % simple URL typesetting
\usepackage{booktabs}       % professional-quality tables
\usepackage{amsfonts}       % blackboard math symbols
\usepackage{nicefrac}       % compact symbols for 1/2, etc.
\usepackage{microtype}      % microtypography
\usepackage{xcolor}         % colors

\title{PolyRecommender: A Multimodal Recommendation System for Polymer Discovery}

\author{
  Xin Wang\thanks{Corresponding authors.}, Yunhao Xiao, Rui Qiao$^{*}$ \\
  Department of Mechanical Engineering\\
  Virginia Tech\\
  Blacksburg, VA 24060 \\
  \texttt{\{xinwang, xyunhao, ruiqiao\}@vt.edu} \\
}

\begin{document}

\maketitle
\setlength{\textfloatsep}{6pt}   % top/bottom floats vs text
\setlength{\intextsep}{6pt}      % “here” floats vs text
\setlength{\floatsep}{-10pt}

\begin{abstract}
  We introduce PolyRecommender, a multimodal discovery framework that integrates chemical language representations from PolyBERT with molecular graph-based representations from a graph encoder. The system first retrieves candidate polymers using language-based similarity and then ranks them using fused multimodal embeddings according to multiple target properties. By leveraging the complementary knowledge encoded in both modalities, PolyRecommender enables efficient retrieval and robust ranking across related polymer properties. Our work establishes a generalizable multimodal paradigm, advancing AI-guided design for the discovery of next-generation polymers.
\end{abstract}

\section{Introduction}
The rational design of novel polymers is a grand challenge in materials science, with the potential to unlock breakthroughs in sustainable energy, advanced manufacturing, and medicine \cite{harun2024emerging}. However, the chemical space of known polymers is astronomically large, making exhaustive experimental screening for specific applications intractable. This creates a critical need for AI-guided discovery frameworks that can intelligently navigate this landscape to recommend candidates with targeted property profiles. The primary bottleneck for such data-driven systems is the development of a polymer representation that is both computationally efficient and chemically informative.

Most prior work adopts unimodal molecular representations. Graph neural networks (GNNs) leverage explicit bond topology and perform well in smaller data regimes but can underrepresent higher-level chemical semantics \cite{gurnani2023polymer, queen2023polymer}. Conversely, transformer models trained on SMILES strings capture chemical "grammar" but may lose critical structural information \cite{weininger1988smiles, kuenneth2023polybert, xu2023transpolymer}. Relying on any single modality provides an incomplete view of a material, limiting a model's ability to generalize and hindering the full potential of AI-guided material design.

\begin{figure}[htbp]
    \centering
    \includegraphics[width=0.9\linewidth]{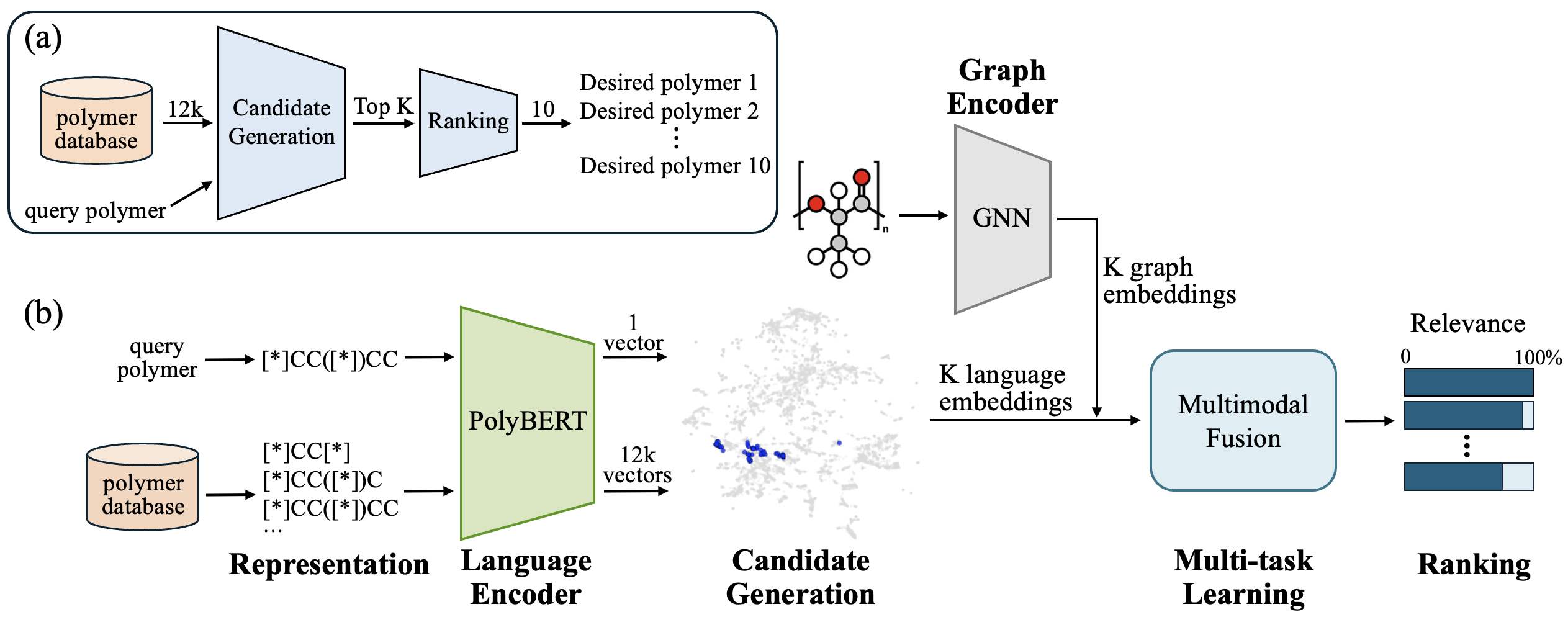}
    \caption{(a) The polymer recommender where polymers are recalled and ranked based on the similarity to the query polymer. (b) The detailed workflow to recommend candidates from the search space including material representation generation, candidate retrieval, fusion of multimodal representations, multi-task prediction for ranking.}
    \label{main}
\end{figure}

To address these limitations and advance the paradigm of polymer design, we introduce PolyRecommender. Our framework operationalizes a two-stage "funnel" architecture \cite{covington2016deep, qu2023language}, a design crucial for practical discovery workflows that require both efficient exploration and precise ranking (Figure~\ref{main}a). The first stage leverages a fine-tuned language model for rapid candidate retrieval from a space of 12,441 polymers, making the initial search computationally tractable. In the second, multimodal ranking stage, we fuse language and graph embeddings to perform a more holistic and accurate evaluation of the top candidates. After systematically investigating three fusion strategies \cite{poria2017review}, our results demonstrate that this multimodal approach consistently outperforms single-modality baselines (Figure~\ref{main}b). This work establishes a powerful and scalable blueprint for next-generation AI-guided discovery systems, effectively integrating chemical language and structural data to accelerate the design of novel polymers.
\section{Related Work}
\label{gen_inst}
The machine learning-driven approach to polymer property prediction has evolved from traditional descriptor-based methods to sophisticated deep learning architectures. The current paradigm is dominated by two powerful, complementary axes of representation learning. First, GNNs directly leverage the molecular graph, enabling models to learn from the explicit topology of atoms and bonds. This approach has proven highly effective, with multitask frameworks demonstrating the capacity to generate robust, generalizable representations \cite{gurnani2023polymer, queen2023polymer}. In parallel, progress in natural language processing has inspired the use of Transformer-based models to treat SMILES strings as a chemical language. Models like PolyBERT \cite{kuenneth2023polybert} and TransPolymer \cite{xu2023transpolymer} learn to encode rich chemical syntax and semantics into dense vector representations. Han et al.\cite{han2024multimodal} established a multimodal transformer fusing semantic and structural embeddings, leading to superior performance in multitask property prediction. Our work operationalizes this validated multimodal synergy within a practical discovery framework, using language representations for efficient candidate retrieval and combined multimodal representations for high-precision ranking.
\section{Methodology}
\subsection{Dataset}
Our experiments are conducted on a dataset of 12,441 synthesized polymers curated from the PolyInfo database \cite{otsuka2011polyinfo}. For each polymer, we use its SMILES representation and three experimental properties: glass-transition temperature ($T_\mathrm{g}$), melting temperature ($T_\mathrm{m}$), and band gap ($E_\mathrm{g}$). The dataset was divided into training, validation, and test sets in a ratio of 80:10:10. We train all multitask models using a masked mean-squared-error (MSE) loss that is computed solely over available ground-truth labels.
\subsection{Dual-modality polymer representations}
To create a comprehensive polymer representation, we employ a dual-modality approach that fuses language embeddings, derived from SMILES strings, with graph embeddings that encode the molecular topology.

\textbf{Language Embeddings.} For our language-based modality, we leverage PolyBERT \cite{kuenneth2023polybert}, a transformer pre-trained on a vast corpus of 100 million polymer SMILES. To adapt this powerful base model for our specific predictive tasks, we employ parameter-efficient fine-tuning using Low-Rank Adaptation (LoRA) \cite{hu2022lora}, which modifies the network's attention layers. The resulting fine-tuned model processes polymer SMILES strings (truncated to 160 tokens) to generate 600-dimensional language embeddings.

\textbf{Graph Embeddings.} To capture structural and topological information, we generate graph embeddings using a Directed Message Passing Neural Network (D-MPNN) \cite{gilmer2017neural, yang2019analyzing}. Each polymer was represented as a molecular graph, where nodes correspond to atoms and edges represent covalent bonds. Node features include atomic number, degree, formal charge, hybridization state, aromaticity, and hydrogen count, while edge features encode bond type and conjugation status. The D-MPNN, architected with 5 message-passing layers and a 512-dimensional hidden state, was trained in a multi-task regression framework to predict three key polymer properties. This pre-training step compels the network to learn chemically meaningful representations, from which the final 512-dimensional graph embeddings are extracted for downstream fusion.
\subsection{Multi-modal fusion}
We investigate three fusion architectures that combine the frozen, pre-computed language ($z^{\text{lang}}$) and graph ($z^{\text{graph}}$) embeddings, training each via multitask regression to predict our three target properties.

\textbf{Early Fusion.}
Following the "shared-bottom" multitask paradigm \cite{caruana1997multitask}, our early fusion model first concatenates the language ($z^{\text{lang}}$) and graph ($z^{\text{graph}}$) embeddings (Eq.~\ref{eq:early_concat}), then processes the resulting vector through a shared 3-layer MLP to produce the 3-dimensional property prediction (Eq.~\ref{eq:ef_network}):
\begin{align}
    x &= [z^{\text{lang}}; z^{\text{graph}}] \label{eq:early_concat} \\
    y &= h(x) \in \mathbb{R}^3 \label{eq:ef_network}
\end{align}
\textbf{Gated Late Fusion.}
To enable modality-specific processing, we implement a gated late fusion model. First, the language ($z^{\text{lang}}$) and graph ($z^{\text{graph}}$) embeddings are independently processed by two dedicated 3-layer MLP “experts” to produce per-task predictions:
\begin{equation}
    y^{\text{lang}} = h^{\text{lang}}(z^{\text{lang}}), \qquad
    y^{\text{graph}} = h^{\text{graph}}(z^{\text{graph}})
    \label{eq:late_fusion_experts}
\end{equation}
A separate gating network (a 2-layer MLP) then takes the concatenated embeddings as input and outputs a task-specific \emph{gating vector} $\mathbf g$. The final prediction of task $k$ is a weighted combination of the expert outputs, dynamically controlled by $g_k$:
\begin{align}
    \mathbf g &= \sigma\!\big(W_g [z^{\text{lang}}; z^{\text{graph}}]\big) \in \mathbb{R}^3 \label{eq:late_fusion_gate} \\
    y_k &= g_k \, y^{\text{lang}}_k \;+\; (1 - g_k)\, y^{\text{graph}}_k \label{eq:late_fusion_final}
\end{align}
\textbf{Multi-gate Mixture-of-Experts (MMoE).}
The MMoE model \cite{ma2018modeling} processes the concatenated input $x=[z^{\text{lang}}; z^{\text{graph}}]$ using $n$ shared experts $\{f_i\}_{i=1}^n$ where $n = 4$. For each task $k$, a dedicated gating network $g^{(k)}(x)$ produces a softmax distribution over experts. A task-specific tower $h^{(k)}$ then maps the gated expert mixture to the final prediction $y_k$:
\begin{align}
    f^{(k)}(x) &= \sum_{i=1}^{n} g^{(k)}_i(x)\, f_i(x) \label{eq:mmoe_mix} \\
    y_k &= h^{(k)}\!\big(f^{(k)}(x)\big) \label{eq:mmoe_tower} \\
    \text{where} \quad g^{(k)}(x) &= \operatorname{softmax}(W^{(k)}_g x) \in \mathbb{R}^n \label{eq:mmoe_gate}
\end{align}

\section{Results}
We developed PolyRecommender, a two-stage multimodal recommendation system designed to efficiently search large chemical spaces for polymers with desired properties. The system employs a "funnel" architecture consisting of two sequential stages: candidate retrieval and multimodal ranking. In the retrieval stage, we use embeddings from a fine-tuned PolyBERT model \cite{kuenneth2023polybert} to represent each polymer. Based on the cosine similarity between these language embeddings, the system retrieves the top 100 candidates most relevant to a given query polymer. In the ranking stage, we refine this list using a multimodal approach that fuses the language embeddings with structural graph embeddings from a GNN. After systematically evaluating three fusion strategies, we selected the MMoE for its superior overall performance in predicting and ranking candidates based on their target properties. The final ranking score is defined in the Appendix.

\begin{figure}[htbp]
    \centering
    \includegraphics[width=0.8\linewidth]{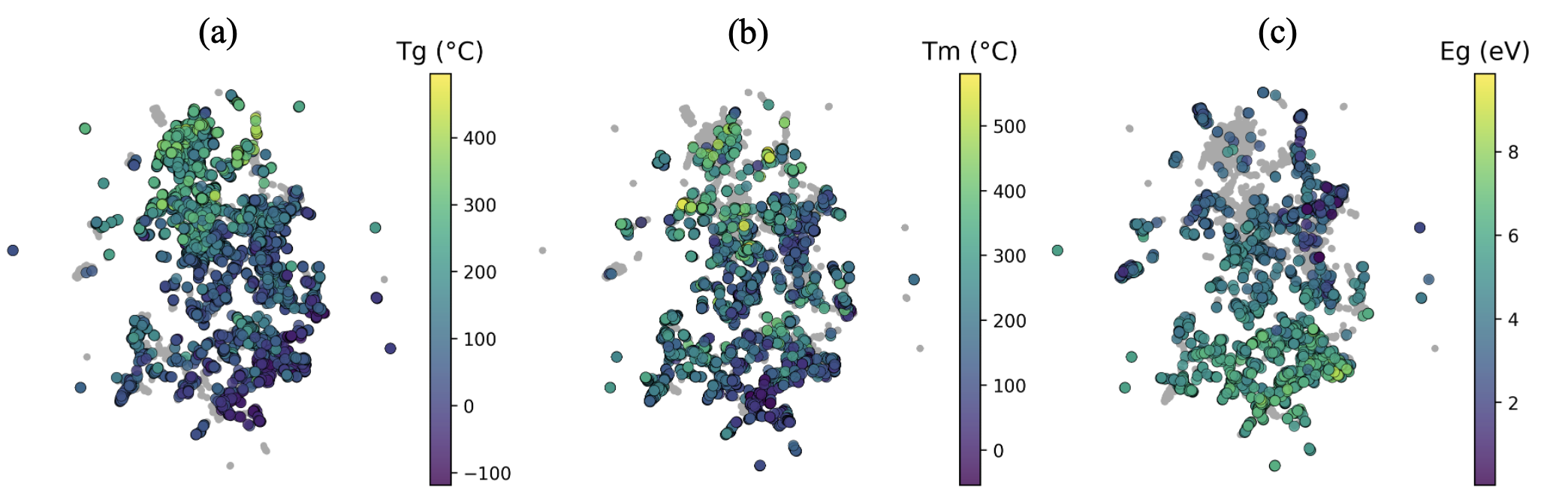}
    \caption{Two-dimensional UMAP projection of the multimodal polymer embeddings. The distribution is colored by three distinct properties: (a) $T_{g}$, (b) $T_{m}$, and (c) $E_g$.}
    \label{fig:umap_distri}
\end{figure}

To validate the quality of our multimodal representations before the final ranking, we visualized the concatenated language and graph embeddings for all 12,441 polymers using a two-dimensional UMAP projection (Figure \ref{fig:umap_distri}). The resulting map reveals distinct clusters corresponding to key polymer properties, which demonstrates that our embeddings successfully capture chemically meaningful relationships and effectively structure the chemical space.

Table~\ref{tab:results} presents a comprehensive ablation study to validate the effectiveness of each component within PolyRecommender, alongside a performance comparison to a state-of-the-art (SOTA) multimodal baseline \cite{han2024multimodal}. Our results show that MMoE fusion achieves the best overall performance among the tested fusion strategies in predicting the glass transition temperature ($T_g$) and band gap ($E_g$). While the gated late fusion model showed a marginal advantage for melting temperature ($T_m$), MMoE provided the best-balanced performance across all tasks. Consistent with prior work \cite{kuenneth2023polybert, han2024multimodal}, we found that predicting $T_m$ is markedly more challenging than predicting $T_g$ or $E_g$. We attribute this difficulty to a potentially weaker correlation between a polymer's melting temperature and its molecular structure.

\begin{figure}[htbp]
    \centering
    \includegraphics[width=0.8\linewidth]{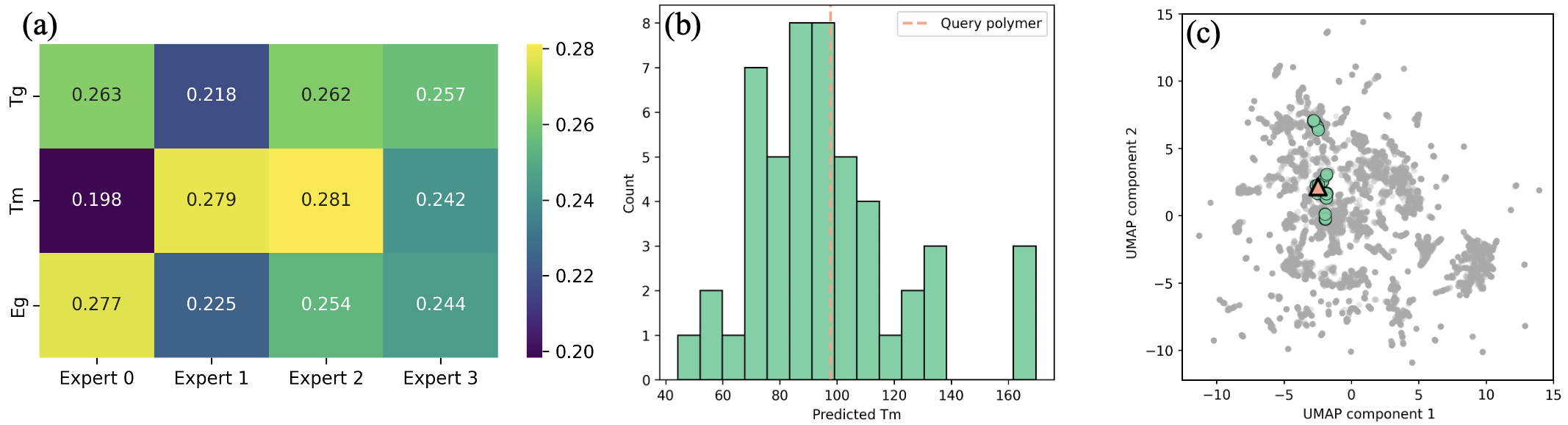}
    \caption{MMoE model analysis: (a) heatmap of task-specific expert utilization; (b) predicted $T_m$ distribution and (c) UMAP projection in space for top 50 candidates from a PEO query.}
    \label{mmoe}
\end{figure}

Our MMoE model demonstrates clear expert specialization (Figure~\ref{mmoe}a), successfully learning task-specific representations for predicting distinct polymer properties ($T_g$, $T_m$, and $E_g$) from a shared input. A case study using Polyethylene oxide (PEO) as a query highlights our system's practical utility. As shown in Figure~\ref{mmoe}b-c, the top 50 recommended candidates not only cluster near PEO in the chemical embedding space but also show a tight distribution of predicted melting temperatures close to the query's known value. This result validates our two-stage "retrieve and rank" framework, confirming it identifies candidates that are both structurally similar and functionally relevant to the user's query.
\begin{table}[ht]
\caption{Test $R^2$ scores for the multi-task prediction. The final model MMoE (Lang + Graph) is compared against a SOTA baseline and several ablation models. The best results are \textbf{bolded}.}
\label{tab:results}
\centering
\begin{tabular}{@{}lccc@{}}
\toprule
\textbf{Model Configuration} & \textbf{$T_\mathrm{g}$} & \textbf{$T_\mathrm{m}$} & \textbf{$E_\mathrm{g}$} \\
\midrule
% --- State-of-the-Art Baseline ---
Multimodal Transformer \cite{han2024multimodal} & 0.880 & 0.720 & -- \\
\midrule
% --- Unimodal Models (Ablation) ---
PolyBERT (Pretrained) & 0.801 & 0.498 & 0.667 \\
GNN (Graph only) & 0.895 & 0.829 & 0.908 \\
PolyBERT (Finetuned) & 0.888 & 0.726 & 0.904 \\
\midrule
% --- Multimodal Fusion Models ---
MMoE (Lang only) & 0.898 & 0.761 & 0.915 \\
Early Fusion (Lang + Graph) & 0.912 & 0.835 & 0.926 \\
Gated Late Fusion (Lang + Graph) & 0.915 & \textbf{0.840} & 0.926 \\
\textbf{MMoE (Lang + Graph)} & \textbf{0.923} & 0.838 & \textbf{0.933} \\
\bottomrule
\end{tabular}
\end{table}

\bibliographystyle{unsrtnat}
\bibliography{references}

%%%%%%%%%%%%%%%%%%%%%%%%%%%%%%%%%%%%%%%%%%%%%%%%%%%%%%%%%%%%
\appendix
\section{Experimental Settings}

\subsection{Hyperparameter Settings}

In this section, we provide the detailed hyperparameter settings used for the models in PolyRecommender. The PolyBERT fine-tuning, GNN training, and the training of the three multimodal fusion models were all conducted on a single NVIDIA A10G GPU. Table~\ref{tab:hyperparams} lists the common training hyperparameters and specific network architectures for each fusion model. The hyperparameters for the GNN and PolyBERT fine-tuning stages largely followed those outlined in their original works.
\begin{table}[h]
\caption{Hyperparameters for the three multimodal fusion models.}
\label{tab:hyperparams}
\centering
\begin{tabular}{@{}ll@{}}
\toprule
\textbf{Hyperparameter} & \textbf{Value} \\
\midrule
Batch Size & 128 \\
Learning Rate & 1e-5 \\
Weight Decay & 1e-3 \\
Total Epochs & 100 \\
Dropout & 0.4 \\
Optimizer & AdamW \\
Learning Rate Scheduler & ReduceLROnPlateau \\
\midrule
\multicolumn{2}{@{}l}{\textit{Architectures}} \\
\midrule
MLP in Early Fusion & 3-layer (hidden sizes [256, 128]) \\
Expert network (Gated Late Fusion) & 3-layer MLP (hidden sizes [256, 128]) \\
Gate network (Gated Late Fusion) & 2-layer MLP (hidden size 128) \\
Number of Experts (MMoE) & 4 \\
Expert network (MMoE) & 3-layer MLP (hidden size 256) \\
Gate network (MMoE) & 2-layer MLP (hidden size 256) \\
Tower network (MMoE) & 2-layer MLP (hidden size 128) \\
\bottomrule
\end{tabular}
\end{table}
\subsection{Dataset Information}

\begin{figure}[h]
    \centering
    \includegraphics[width=0.9\linewidth]{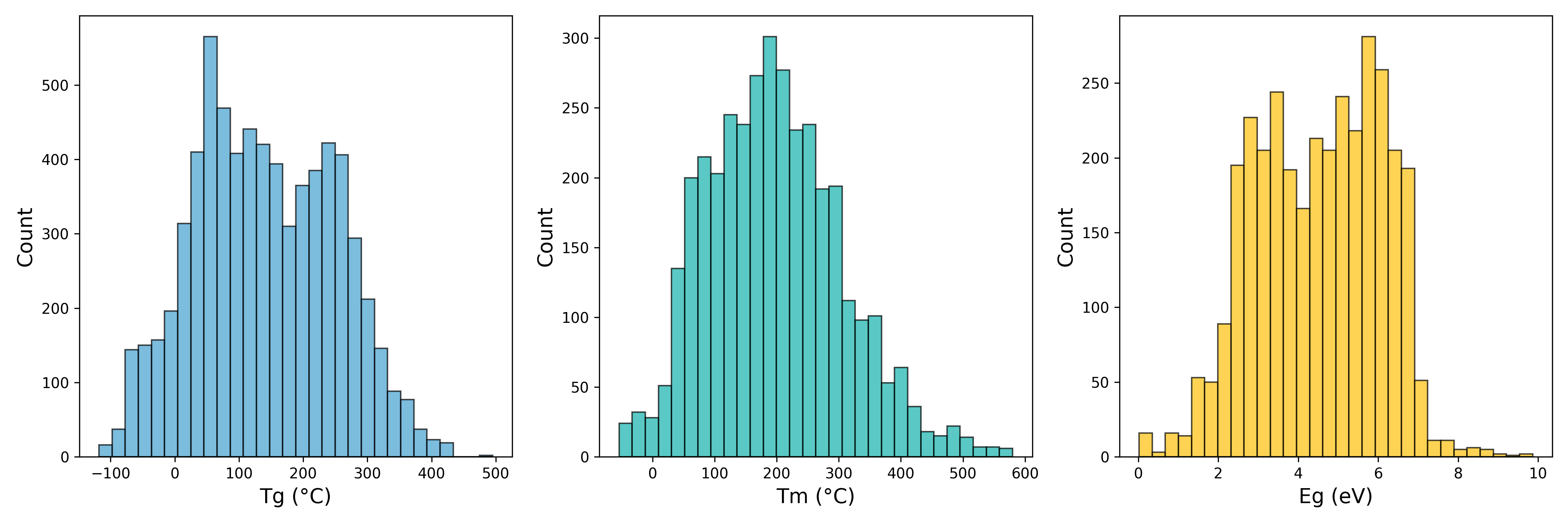}
    \caption{Distributions of available experimental data for the three target properties.}
    \label{fig:histo}
\end{figure}

The dataset used in this work consists of 12,441 synthesized polymers curated from the PolyInfo database. The distributions of the available experimental data for the three target properties are shown in Figure~\ref{fig:histo}. The dataset exhibits a significant label imbalance, with many more values available for glass transition temperature ($T_\mathrm{g}$, 6900 labels) than for melting temperature ($T_\mathrm{m}$, 3633 labels) and band gap ($E_\mathrm{g}$, 3379 labels).

\subsection{Relevance Score}

To rank the final list of candidates for a given query, we defined a unitless relevance score, $R$ (from 0 to 100), based on the Total Absolute Percentage Difference (TAPD) across all shared properties:
\begin{align}
    R = \frac{100}{\text{TAPD} + 1}, \quad
    \text{TAPD} = \sum_{i=1}^{n} \left| \frac{y_i^{c} - y_i^{q}}{y_i^{q}} \right|
\end{align}
where $y_i^c$ and $y_i^q$ are the predicted values of the $i$-th property for the candidate and query polymer, respectively, and $n$ is the number of properties being compared. This metric normalizes the error across properties of different scales, and the $+1$ in the denominator ensures a perfect match (TAPD=0) yields a maximum score of 100.

\end{document}